\documentclass{spie}

\usepackage[utf8]{inputenc}

\usepackage[table,xcdraw]{xcolor}
\usepackage{times} 
\usepackage{indentfirst} 
\usepackage{multirow}
\usepackage{graphicx}
\usepackage{xkeyval}
\usepackage{adjustbox}
\usepackage{booktabs, multirow} 
\usepackage{soul}
\usepackage[table]{xcolor} 
\usepackage{changepage,threeparttable} 
\usepackage{amsmath,graphicx, amsfonts}
\usepackage{tabularx}
\usepackage{booktabs, multirow} 
\usepackage{soul}
\usepackage[table]{xcolor} 
\usepackage{changepage,threeparttable}
\usepackage{algorithm}
\usepackage{algorithmic}
\usepackage{amsmath}
\usepackage{hyperref}

\newcommand\doubleplus{\mathbin{+\mkern-10mu+}}

\title{Generalization of Brady-Yong Algorithm for Fast Hough Transform to Arbitrary Image Size}

\author{Danil Kazimirov\supit{1,2,3},
Dmitry Nikolaev\supit{1,4},
Ekaterina Rybakova\supit{1,3},
Arseniy Terekhin\supit{2}
\skiplinehalf
  \normalsize
\supit{1} Smart Engines Service LLC, 117312 Moscow, Russia; \\
\supit{2} Institute for Information Transmission Problems (Kharkevich Institute) RAS, 127051 Moscow, Russia; \\
\supit{3} Faculty of Mechanics and Mathematics, Lomonosov Moscow State University, 119991 Moscow, Russia; \\
\supit{4} Federal Research Center Computer Science and Control RAS, 119333 Moscow, Russia.
}

\usepackage{mathptmx}
\begin{document}

\maketitle

\begin{abstract}

Nowadays, the Hough (discrete Radon) transform (HT/DRT) has proved to be an extremely powerful and widespread tool harnessed in a number of application areas, ranging from general image processing to X-ray computed tomography.
Efficient utilization of the HT to solve applied problems demands its acceleration and increased accuracy.
Along with this, most fast algorithms for computing the HT, especially the pioneering Brady-Yong algorithm, operate on power-of-two size input images and are not adapted for arbitrary size images.
This paper presents a new algorithm for calculating the HT for images of arbitrary size.
It generalizes the Brady-Yong algorithm from which it inherits the optimal computational complexity.
Moreover, the algorithm allows to compute the HT with considerably higher accuracy compared to the existing algorithm.
Herewith, the paper provides a theoretical analysis of the computational complexity and accuracy of the proposed algorithm.
The conclusions of the performed experiments conform with the theoretical results.


\end{abstract}

\section{Introduction}
\label{sec:introduction}

The Hough transform (HT) is a widely used tool in image processing and machine vision.
It is usually understood as a method for robust estimation of the parameters of one or more lines in a discrete image by calculating the number of points lying on each line of a set of discrete lines.
The method is named after Paul Hough, who first proposed it in 1959 as a way to identify straight-line tracks in a bubble chamber~\cite{Hough(1959)}.
The core idea of the HT is to count ``votes'' along discrete, parameterized lines, and accumulate a value for each of them. The higher this value is, the greater the probability that the line with corresponding parameter values is present in the image.
Even though the best-known application of the HT is to find contrasting straight lines or their segments in an image~\cite{Illingworth(1988), Chaloeivoot(2016)}, over the years, the HT has found numerous applications in a variety of fields, including image binarization~\cite{Aliev(2014)}, segmentation~\cite{Ozturk(2018)}, computed tomography~\cite{Polevoy(2023)}, etc.

The HT requires fast algorithms for its calculation.
In 1992, M. Brady and W. Yong published a paper on the fast HT/DRT~\cite{Brady(1992)} for images of size $n \times n, \, n = 2^q, \, q \in \mathbb{N}$.
The method they proposed is based on dynamic programming, which is used to skip the repeated calculation of the sums of already processed segments when computing the sum along the line they constitute.
This allows for the calculation of the HT in $\Theta(n^2 \log_2 n)$ summations.
This method has become the de facto standard for practical applications of the Hough transform, as it offers a computationally efficient approach while maintaining the core functionality of the original method.
This allowed the HT to be used in systems that have strict hardware constraints with a requirement to perform in real-time~\cite{Arlazarov(2022), Kotov(2015)}.
In this paper, we propose a novel fast HT computation algorithm generalizing the Brady-Yong algorithm to arbitrary image size.
The proposed algorithm, described in section~\ref{sec:algorithm_description}, has the same complexity as the Brady-Yong algorithm while being more accurate than the previously known algorithm of the same kind~\cite{Anikeev(2021)}.
In section~\ref{sec:complexity_and_accuracy_theoretical_analysis}, we analyze the computational complexity and accuracy of the proposed algorithm, which align with the experimentally derived estimates outlined in section \ref{sec:experiments} of the paper.

\section{Related works}
\label{sec:related_works}

The history of the HT originated in 1959~\cite{Hough(1959)} and since that time many papers have clarified and developed the idea of the HT.
Nowadays it is acknowledged that, mathematically, the HT is a discrete form of the Radon transform~\cite{Illingworth(1988)}.
The fastest known algorithm to compute HT/DRT for now is the Brady-Yong algorithm~\cite{Brady(1992)}.
The authors used dyadic patterns to approximate straight lines in the image.
Their algorithm relies on the repeated use of common subsums and computes the HT for a square image of size $n \times n$ in $\Theta(n^2 \log_2 n)$ summations, where $n = 2^q, q \in \mathbb{N}$.
Addressing the accuracy of the algorithm, S. Karpenko and E. Ershov, in 2021~\cite{Karpenko(2021)}, estimated the maximal deviation between a dyadic pattern and a geometric straight line and showed that if image linear size is $2^q,\, q \in \mathbb{N}$, the maximal orthotropic error of approximating geometric lines amounts to exactly $\log_2 n / 6$ for even $q$ and $\log_2 n / 6 - 1 / 18$ for odd $q$.

A significant limitation of the Brady-Yong algorithm is that it is applicable only for images with sides equal to the powers of two.
To overcome this constraint, researchers have described several unifications applicable to images of arbitrary size.
In 2012, J. Marichal-Hernandez \emph{et al.} proposed the modification~\cite{Marichal-Hernandez(2012)} based on the decomposition of image sizes into multipliers.
By its design, the algorithm is inefficient in the case of decomposition containing large prime factors.
A universal generalization of the Brady-Yong algorithm was published in 2021 by F. Anikeev with co-authors~\cite{Anikeev(2021)}.
This variant preserves the scheme and asymptotic complexity of the Brady-Yong algorithm but the accuracy of this algorithm was not estimated. 

In 2018, T. Khanipov proved~\cite{Khanipov(2018)} the $\Omega (n^2 \log_2 n)$ asymptotic lower bound on a number of operations for computing the HT using dyadic patterns.
It follows that the Brady-Yong algorithm is non-improvable in terms of speed and can be regarded as a baseline for generalization in application to images with arbitrary linear size.
Further in the paper, we provide a detailed description of the known and novel algorithms based on the idea of Brady and Yong.
These algorithms maintain the same computational complexity as the original one but are suitable for images of arbitrary size.

\section{Algorithm description}
\label{sec:algorithm_description}

In the current section, we describe the proposed fast HT algorithm $FHT2DT$ (Fast HT by the Dichotomy, Tweaked).
Parallelly, we recall the known algorithm~\cite{Anikeev(2021)} that is valid for arbitrary image size $n \in \mathbb{N}$, which we would refer to as $FHT2DS$ (Fast HT by the Dichotomy, Simple).
The algorithms are quite similar, so it is convenient to introduce the $FHT2D$ algorithm (Algorithm~\ref{alg:FHT2D}) as a template for $FHT2DS$ and $FHT2DT$ algorithms.
Here, to address subranges of the vectors, we adhere to slicing notation, i.e. $n_1 : n_2$ indicates the range from $n_1$ to $n_2$ (including $n_1$, not including $n_2$).
The absence of both indices indicates the full range. 

\begin{algorithm}[H]
\caption{$FHT2D$}\label{alg:FHT2D}
    \begin{algorithmic}[1]
    \STATE{\textbf{Input:} $w > 0,\, h > 0,\,$ image $I_{w \times h}$, function $Split \in \{ Split_S, Split_T \}$}
    \STATE{\textbf{Output:} Hough image $J_{w \times h}$}
    \IF{$w > 1$}
        \STATE{$w_P \leftarrow Split(w)$} \COMMENT{$w_P$ is a pair of numbers}
        \STATE{$I_P \leftarrow \langle I(0 : w_P(0),\, :),\, I(w_P(0) : w,\, :)\rangle$} \COMMENT{$I_P$ is a pair of images}
        \STATE{$J_P \leftarrow FHT2D(w_P, \langle h, h \rangle, I_P, \langle Split, \, Split \rangle)$} \COMMENT{A pair of $FHT2D$ calls builds image pair $J_P$}
        \STATE{$J \leftarrow Create\_Zeroed\_Image(w, h)$} \COMMENT{$J$ is an image}
        \STATE{$k_P \leftarrow (w_P - 1) / (w - 1)$} \COMMENT{$k_P$ is a pair of numbers}
        \FOR{$t \leftarrow 0$ \textbf{to} $w - 1$}
            \STATE{$R_P \leftarrow J_P([t \, k_P], \, :)$} \COMMENT{$R_P$ is a pair of vectors}
            \STATE{$J(t,\, :) \leftarrow R_P(0) + Rotate\left(R_P(1), (t - [t \, k_P(1)]) \mod h\right)$}
        \ENDFOR
    \ELSE
        \STATE{$J \leftarrow I$}
    \ENDIF
    \end{algorithmic}
\end{algorithm}

The $FHT2D$ template algorithm consists of two stages.
Suppose that we have an image $I$ of size $w \times h$.
The first stage is a recursive image $I$ splitting into left $I_P(0)$ and right $I_P(1)$ parts.
The left part gets size equal to $w_P(0) \times h$ and the right part size equals $w_P(1) \times h$.
Here, $w_P$ stands for a result of the input $Split$ function call for the input image width $w$ (see line 4 in Algorithm \ref{alg:FHT2D}).
The input $Split$ function definition is the only place that distinguishes two algorithms.
Within the previously known $FHT2DS$ algorithm, in application to the width $w$, the input $Split$ function returns tuple $Split_S(w) = \langle \lfloor w/2 \rfloor, \lceil w/2 \rceil \rangle$, whereas within the proposed $FHT2DT$ algorithm the $Split$ function returns tuple $Split_T(w) = \langle 2^{\lceil \log_2 w  \rceil - 1}, w - 2^{\lceil \log_2 w  \rceil - 1} \rangle$.
In other words, $FHT2DS$ bisects $w$ while $FHT2DT$ cuts off the maximal power-of-two.

The second stage merges the tuple $J_P$ of the recursively computed HT of the parts $I_P$ until the desired Hough image is obtained.
The merging is implemented by repeated execution of the same vector operation on different data. This operation (Algorithm \ref{alg:FHT2D}, line 11) consists in summing two vectors one of which is circularly shifted.
Herewith, $Rotate(v, r)$ denotes the circular shifting of the vector $v$ elements by $r$ positions.
The summation results are put into the image $J$ which is initialized by zeros via the $\emph{Create\_Zeroed\_Image(w, h)}$ function.

Note that when $n=2^q$, $q \in \mathbb{N}$, $FHT2DS$ and $FHT2DT$ algorithms coincide with the Brady-Yong algorithm~\cite{Brady(1992)}.
So, both of them might be regarded as a generalization of the Brady-Yong algorithm to arbitrary image size.

$FHT2D$ algorithm calculates sums of input image $I$ values over discrete patterns that approximate straight lines within an image region.
All such patterns can be obtained by vertically shifting the generating patterns, which approximate predominantly horizontal geometric lines, i.e. lines of the form $y = x \, t / (n - 1)$, $t \in \mathbb{Z}_n$~\cite{Karpenko(2021)}.
The sets of generating patterns $\mathcal{P}_S(n)$ and $\mathcal{P}_T(n)$ along which an image of size $n \times n$ is integrated using $FHT2DS$ and $FHT2DT$ algorithms are directly determined by the certain input $Split$ function definition.
Function constructing generating patterns with known $n$ and $t$ in form $y = pat(x)$, $x, y \in \mathbb{Z}_n$, for $FHT2D$ algorithm is shown in Algorithm \ref{alg:build-fht2d-pattern}.
Here, $a \doubleplus b$ refers to the concatenation of the tuples $a$ and $b$.

\begin{algorithm}[H]
  \caption{$FHT2D\_Pattern$}\label{alg:build-fht2d-pattern}
  \begin{algorithmic}[1]
    \STATE{\textbf{Input:} $n > 0$, $0 \leq t < n$, function $Split \in \{ Split_S, Split_T \} $}
    \STATE{\textbf{Output:} Generating pattern $pat$}
    \IF{$n > 1$}
      \STATE $n_P \gets Split(n)$ \COMMENT{$n_P$ is a pair of numbers}
      \STATE $t_P \gets \left[t \, (n_P - 1) / (n - 1)\right]$ \COMMENT{$t_P$ is a pair of numbers}
      \STATE $pat_P \gets FHT2D\_Pattern(n_P, t_P, \langle Split, \, Split \rangle ) + \langle 0, t - t_P(1) \rangle$ \COMMENT{A pair of $FHT2D\_Pattern$ calls builds pair $pat_P$}
      \STATE $pat \gets pat_P(0) \doubleplus pat_P(1)$ \COMMENT{$pat$ is a tuple}
    \ELSE
      \STATE $pat \gets \langle 0 \rangle$
    \ENDIF
  \end{algorithmic}
\end{algorithm}

\section{Theoretical analysis of complexity and accuracy}
\label{sec:complexity_and_accuracy_theoretical_analysis}

Denote by $f_{S}(n)$ and $f_{T}(n)$ the number of summations required by the $FHT2DS$ and $FHT2DT$ algorithms respectively while processing square image $I$ of size $n \times n$, $n \in \mathbb{N}$.

\textbf{Theorem 1. The following sharp estimations for $f_{S}(n)$ and $f_{T}(n)$ hold
\begin{gather}\label{eq:complexity_fht2ds}
  f_{S}(n) = (\lfloor \log_2 n \rfloor + 2) \, n^2 - 2^{\lfloor \log_2 n \rfloor + 1} \, n \leq \frac{5 \, \log_3 2}{3} n^2 \log_2 n < 1.052 \, n^2 \log_2 n, \\ \label{eq:complexity_fht2dt}
  f_{T}(n) \leq \frac{81 \, \log_{17} 2}{17} n^2 \log_2 n < 1.166 \, n^2 \log_2 n.
\end{gather}
}

\textit{Proof.}
The recursive call of the $Split$ function generates oriented binary tree graph $G(n)$ which controls the HT calculation order.
Each tree node is associated with the image $I$ part of some size $m \times n$ being processed in the corresponding recursion step.
Let us attribute width $m$ to the corresponding tree node.
Its children nodes possess values $Split(m)(0)$ and $Split(m)(1)$.
According to the $FHT2D$ template, recursion step with image width $m$ requires $m \, n$ summations.
Thus, total summation number $f(n)$ is equal to the sum of values attributed to non-leaf nodes of $G(n)$ multiplied by image height $n$.

Consider graph $G_{S}(n)$ corresponding to the $FHT2DS$ algorithm.
Its depth equals $\lceil \log_2 n \rceil + 1$ and the quantity of its completely filled levels is equal to $\lfloor \log_2 n \rfloor + 1$.
The sum of node values in each completely filled level is $n$.
Thus, node values from all except the last completely filled levels sum up to the value $n \, \lfloor \log_2 n \rfloor$.
Now, let us take into account the non-leaf nodes from the last completely filled tree level (they have values 2): by induction on $n$, the number of values 2 in such level is proved to equal $n-2^{\lfloor \log_2 n \rfloor}$.
In sum, this gives $2 \, (n-2^{\lfloor \log_2 n \rfloor})$.
Hence, we conclude that $f_{S}(n) = n^2 \, \lfloor \log_2 n \rfloor + 2 \, n \, (n - 2^{\lfloor \log_2 n \rfloor}) = (\lfloor \log_2 n \rfloor + 2) \, n^2 - 2^{\lfloor \log_2 n \rfloor + 1} \, n$.

The normalized function $f_{S}(n) / n^2 \log_2 n$ attains its maximum value as soon as $n = 3$ ($f_{S}(n)$ becomes monotonic after several initial values $n$), and the maximum value derives to $5 \log_3 2 / 3 < 1.052$.
This completes the proof of Equation~\eqref{eq:complexity_fht2ds}.

Regarding the tree $G_{T}(n)$ of the proposed algorithm, note that the number of completely filled levels changes non-monotonically whereas for $G_{S}(n)$ this number presents a monotonic sequence.
For the majority values of $n$, the graph $G_{T}(n)$ is imbalanced.
The critical case is when $n= 2^k + 1$, $k \in \mathbb{N}$, i.e. one of the root children subtree has a depth of exactly 1.
In other words, the maximal value of the normalized complexity $f_{T}(n) / (n^2 \log_2 n)$ over the segment $[2^k, 2^{k+1}]$ is attained when $n = 2^k + 1$.
In such case, we calculate $f_{T}(2^k + 1) = (2^k + 1)^2 + (2^k + 1) \, f_{T}(2^k) / 2^k = (2^k + 1)(2^k + 1 + k \, 2^k)$.
Via the derivative analysis, one proves that the normalized complexity $f_{T}(n) / (n^2 \log_2 n) \leq (2^k + 1 + k \, 2^k) / ((2^k + 1) \log_2 (2^k + 1))$ is majorized by the fraction $(81 \, \log_{17} 2) / 17 < 1.166$ that is reached as soon as $k=4$, $n=17$.
$\square$

Theorem 1 implies that the computational complexity of both $FHT2DS$ and $FHT2DT$ algorithms is $\Theta(n^2 \log_2 n)$, and $f_{S}(n), f_{T}(n) \rightarrow n^2 \log_2 n$ as $n \rightarrow \infty$.
This coincides with the behaviour of the Brady-Yong algorithm when $n=2^q$, $q\in \mathbb{N}$.

Next, to investigate the proposed $FHT2DT$ algorithm accuracy, we will estimate the maximal orthotropic error $E_T(n)$ of ideal lines $l(n, t) = \{y = x \, t / (n - 1) \, \big| \, x \in \mathbb{Z}_n \}$ approximation by the patterns $pat(n, t) \in \mathcal{P}_T$ with slope $t \in \mathbb{Z}_n$: $E_T(n) = \max \big\{ \| l(n,t) - pat(n,t) \| \, \big| \, t \in \mathbb{Z}_n, \, pat(n,t) \in \mathcal{P}_{T} \big\}$ where $\| l(n,t) - pat(n,t) \| = \max \big\{ | l(n,t)(x) - pat(n,t)(x) | \, \big| \, x \in \mathbb{Z}_n \big\}$. 

\textbf{Theorem 2. The following inequality for the maximal orthotropic error $E_T(n)$ holds 
\begin{gather} 
    \label{eq:accuracy_fht2dt}
    E_T(n) \leq \frac{\lfloor \log_2 n \rfloor}{6} + 1 - 2^{\displaystyle -\lfloor \log_2 n \rfloor}.
\end{gather}
}

\textit{Proof.} 
Let the square image $I$ size $n$ be equal to $2^{q_0} + 2^{q_1} + \ldots + 2^{q_{N - 1}}$, $0 \leq q_0 < q_1 < \ldots < q_{N - 1}$, $q_i \in \mathbb{Z}$, $i \in \mathbb{Z}_N$, $N \in \mathbb{N}$.
Here, $q_{N - 1} = \lfloor \log_2 n \rfloor$  and $N = ones(n) \leq \lfloor \log_2 n \rfloor + 1$ is a Hamming weight of $n$.
For the fixed values of $n$ and $t$, we appeal to the simplified notations $pat = pat(n, t)$ and $l = l(n, t)$.
Let us associate string $S = s_0 \doubleplus s_1 \doubleplus \ldots \doubleplus s_{k-1}$, $s_i \in \{ L, \, R \}$, $i \in \mathbb{Z}_k$ to each node of the calculation order tree $G_T$, concatenating left $(L)$ or right $(R)$ moves on the path from the root to the considered node.
Let $\overline{S} \subseteq \mathbb{Z}_n$ be a segment of the image along axis $x$ processed in node $S$.
We define $\|l - pat\|_{\overline{S}} = \max \big\{ | l(x) - pat(x) | \, \big| \, x \in \overline{S} \big\}$ to consider the deviation of the pattern $pat$ from the line $l$ within the range $\overline{S}$.

Let us introduce the following notations: $t_L =\left[ t \, \left( 2^{q_{N-1}}-1 \right) / \left( n-1\right) \right]$, $t_R = \left[ t \, 2^{q_N} / \left( n-1 \right) \right]$, $l_0$ is a line connecting points $(0, \,0)$ and $(2^{q_{N-1}}-1, \, t_L)$, $l_R$ is a line connecting points $(2^{q_{N-1}}, \, t_R)$ and $(n-1, \, t)$, $t_{RL}=\left[ l_R(2^{q_{N-1}}+2^{q_{N-2}}-1) \right]$ and $l_1$ stands for a line connecting points $(2^{q_{N-1}}, \, t_R)$ and $(2^{q_{N-1}}+2^{q_{N-2}}-1, \, t_{RL})$. 
Lines $l_0, l, l_R$ and $l_1$ create continuous polygonal chain in $I$. 
Note that $\| l-l_0 \|_{\overline{L}} \leq 1 / 2$, $\| l_R - l_1 \|_{\overline{RL}} \leq 1 / 2$. 

Also, notice that $|l(t_{RL})-l_R(t_{RL})| \leq |l(t_{R})-l_R(t_{R})|/2 \leq 1/4$ due to the fact that $l_R$ and $l$ intersect at point $(n-1, t)$ and length of $\overline{RR}$ is no bigger than length of $\overline{RL}$.
From the latter inequality we conclude that $|l(t_{RL})-l_1(t_{RL})|\leq  |l_R(t_{RL})-l_1(t_{RL})|+|l(t_{RL})-l_R(t_{RL})| \leq 1/2 + 1/4$. 
Therefore, recalling that $|l(t_R)-l_1(t_R)| \leq 1/2$, we derive $\|l-l_1 \|_{\overline{RL}} \leq 1/2+1/4$.

Utilizing triangle inequality for the norm, estimation $\|l-l_0 \|_{\overline{L}} \leq 1/2$ implies that $\|l-pat \|_{\overline{L}} \leq \|l_0 - pat\|_{\overline{L}} + 1/2$. Similarly, estimation $\|l-l_1 \|_{\overline{RL}} \leq 1/2+1/4$ gives that $\|l-pat \|_{\overline{RL}} \leq \|l_1 - pat\|_{\overline{RL}} + 1/2 + 1/4$. 
Uniting the latter inequalities with the estimations~\cite{Karpenko(2021)} $\|l_0 - pat \|_{\overline{L}} \leq E_T(2^{q_{N-1}}) \leq q_{N-1}/6$, $\|l_1 - pat \|_{\overline{RL}} \leq E_T(2^{q_{N-2}}) \leq q_{N-2}/6$, we conclude that $\|l - pat \| \leq \max \left\{ q_{N-1} / 6 + 1 / 2, \: q_{N-2} / 6 + 1 / 2 + 1 / 4, \: \|l - pat \|_{\overline{RR}}  \right\}$.

Continuing analogous inequalities over segments $\overline{RRL}$, $\overline{RRRL}$ and so on, we obtain
\begin{equation} \label{eq:final_ineq}
\|l- pat \| \leq \max_{k \in \mathbb{Z}_N} \left\{ \frac{q_{N-k}}{6} + \sum_{m=1}^{k} \frac{1}{2^m} \right\} \leq \frac{q_{N-1}}{6} + 1 - 2^{-(N-1)} \leq \frac{\lfloor \log_2 n \rfloor}{6} + 1 - 2^{\displaystyle -\lfloor \log_2 n \rfloor}. 
\end{equation}
Taking the supremum over $t$ in the left side of the inequality (\ref{eq:final_ineq}) ends the proof. $\square$

Note, $E_T(n) / \left(\log_2 n /6\right) \rightarrow 1$ as $n \rightarrow \infty$, so $FHT2DT$ has the same asymptotic accuracy as the Brady-Yong algorithm.

\section{Experiments and discussion}
\label{sec:experiments}

\begin{figure}[!ht]
  \centering
  \includegraphics[scale=0.55]{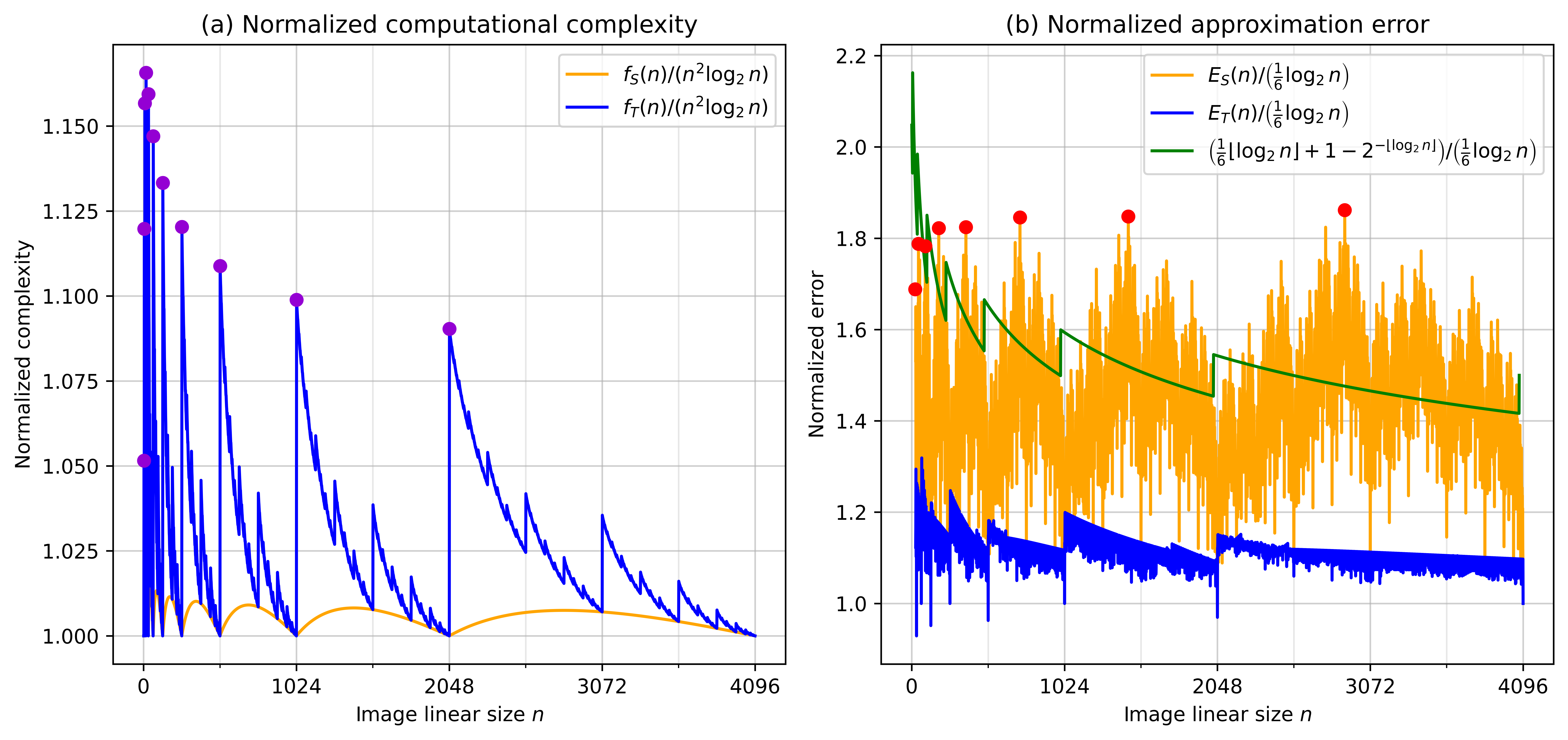}
  \caption{\label{fig:complexity_and_error}
    (a) Experimentally assessed values of the $FHT2DS$ and $FHT2DT$ algorithms computational complexities normalized by $n^2 \log_2 n$. Dark violet points depict peak values of the $FHT2DT$ complexity.
    (b) Maximum orthotropic error of approximation of straight lines by $FHT2DS$ and $FHT2DT$ patterns normalized by $\log_2 n / 6$. Red points show the location of the highest local maxima of the normalized approximation error provided by $FHT2DS$ patterns.
    The green graph highlights the position of the derived in Theorem 2 estimation normalized by $\log_2 n / 6$.
  }
\end{figure}

$FHT2DS$ and $FHT2DT$ algorithms complexities have been computed according to Algorithm~\ref{alg:FHT2D}.
Figure~\ref{fig:complexity_and_error} (a) shows the graphs for numbers of summations $f_S(n)$ and $f_T(n)$ normalized by $n^2 \log_2 n$, as discrete functions of image linear size $n \leq 4096$.

Following Algorithm~\ref{alg:build-fht2d-pattern} to build $FHT2DS$ and $FHT2DT$ patterns and looping over whole sets of patterns $\mathcal{P}_S(n)$ and $\mathcal{P}_T(n)$ for all $n \leq 4096$, we have also evaluated the maximum orthotropic errors $E_S(n)$ and $E_T(n)$ of approximation of the corresponding ideal geometric lines by these sets of patterns (where the quantity $E_S(n)$ is defined analogously to $E_T(n)$). Figure~\ref{fig:complexity_and_error} (b) depicts how the maximum orthotropic approximation errors normalized by $\log_2 n / 6$ change as image linear size $n$ increases. 

\begin{figure}[!ht]
\centering
\includegraphics[scale=0.55]{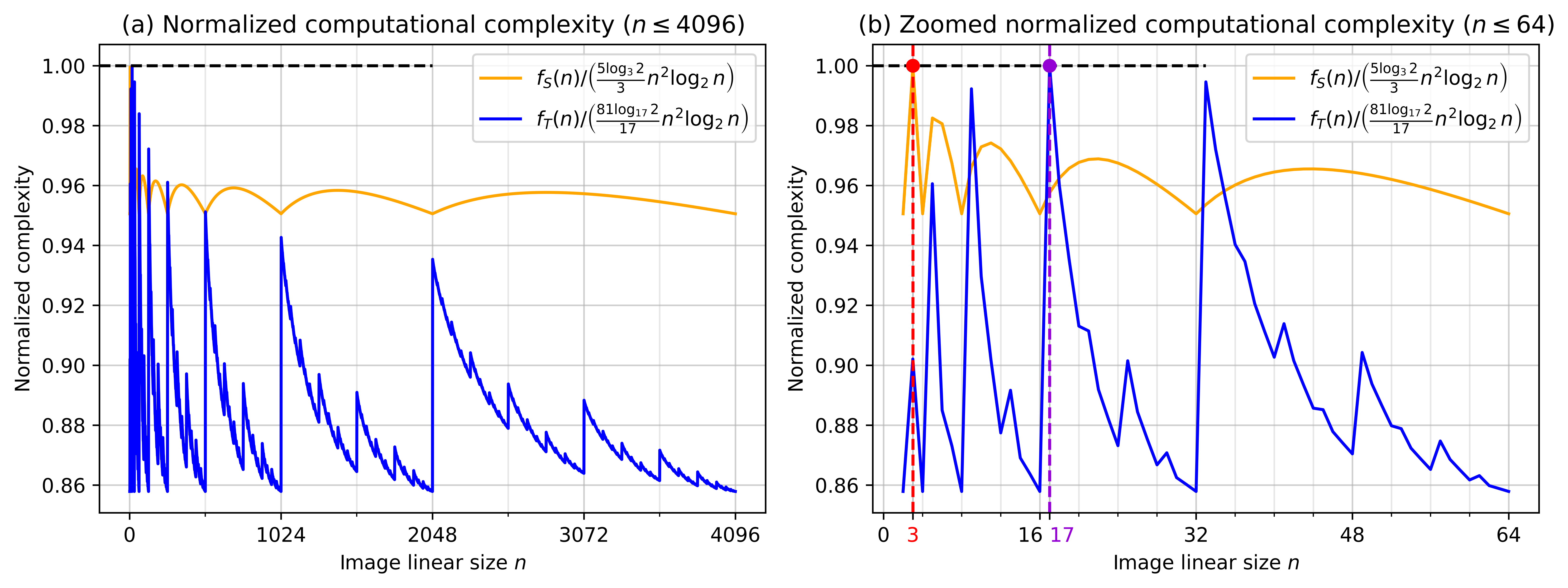}
\caption{\label{fig:complexity_and_zoomed_complexity} Measured values of the computational complexities of $FHT2DS$ and $FHT2DT$ algorithms divided by the corresponding estimating expressions in Theorem 1 (for different image sizes $n$). (a) The dynamics of individually normalized complexities over the full range $n \leq 4096$. (b) Normalized complexity values dynamics over the segment $n \leq 64$. Points and dashed lines mark image sizes for which algorithms normalized complexities equal 1.}
\end{figure}

Figure~\ref{fig:complexity_and_zoomed_complexity} presents graphs of $f_{S}(n) / \left(5 \, (\log_3 2) \, n^2 \log_2 n / 3\right)$ and $f_{T}(n) / \left(81 \, (\log_{17} 2) \, n^2 \log_2 n / 17\right)$ computational complexities normalized by estimations provided by Theorem 1. 
The fact that graphs of the normalized quantities are bounded by unity over the displayed region for $n$ suggests that $f_{S}(n)$ and $f_{T}(n)$ possess the Brady-Yong algorithm computational complexity asymptotics $\Theta(n^2 \log_2 n)$ as $n \rightarrow \infty$ (Figure \ref{fig:complexity_and_zoomed_complexity} (a)). 
As for $n=3$ and $n=17$ normalized complexities take value 1 (see the zoomed region, Figure 2 (b)), the constants in Theorem 1 are sharp and non-improvable among estimations of the form $const \, n^2 \log_2 n$. 
Additionally, the constant in the asymptotic complexity of $FHT2DT$ is larger than that of the $FHT2DS$ algorithm.
Thus, the $FHT2DT$ algorithm slightly degrades the speed of the known $FHT2DS$ algorithm. The difference in speed becomes rather indistinguishable as $n$ increases.

Besides, Figure \ref{fig:complexity_and_error} (b) compares the maximum orthotropic approximation errors $E_S(n)$ and $E_T(n)$ to the estimation derived in Theorem 2 (normalized quantities are visualized in Figure \ref{fig:complexity_and_error} (b)). 
The graph of the estimation $\lfloor \log_2 n \rfloor / 6 + 1 - 2^{- \lfloor \log_2 n \rfloor}$ not only majorizes error $E_T(n)$ of the $FHT2DT$ algorithm, but it also efficiently separates $FHT2DS$ errors from below for no less than 36.38 \% of $n$ within segment $[1, 4096]$. 
For certain image sizes $n$, the maximum orthotropic approximation error $E_S(n)$ appears to be more than 1.69 times larger (namely, for $n=1451$) than the error $E_T(n)$ afforded by the proposed method $FHT2DT$. 
Presumably, the highest peaks of the graph $E_{S}(n) / (\log_2 n / 6)$ correspond to values $n_k$ such that $n_{k+1}=2 \, n_k + (-1)^{k+1}$, $n_0=23$, creating sequence $23, 45, 91, 181, 363, 725, 1451$ and so on. 
To sum up, as algorithms $FHT2DS$ and $FHT2DT$ realize approximation of integrals over straight lines within an image region, all of the aforementioned points allow us to conclude that the proposed $FHT2DT$ algorithm is significantly more accurate than the $FHT2DS$ algorithm.

\section{Conclusion}
\label{sec:conclusion}

The paper proposes a novel $FHT2DT$ algorithm as a generalization of the Brady-Yong algorithm for calculating the HT of images of non-power-of-two linear size. 
The algorithm preserves the optimal computational complexity of the Brady-Yong algorithm. 
Moreover, in comparison with the existing $FHT2DS$ algorithm~\cite{Anikeev(2021)}, the developed algorithm proves to be much more accurate. 
Characteristics such as speed and accuracy of the proposed $FHT2DT$ algorithm are theoretically analyzed. 
Theoretical considerations align with the experimental results. 
The practical potential of the proposed algorithm lies in fields where both accuracy and speed of the image processing matters, such as industrial computer vision systems on edge devices.

\bibliographystyle{spiebib}
\bibliography{main}

\end{document}